\newif\ifieee
\newif\ifral

\ieeetrue
\raltrue

\ifieee
\documentclass[letterpaper, 10 pt, conference]{ieeeconf}  

\makeatletter
\let\NAT@parse\undefined
\makeatother
    
\IEEEoverridecommandlockouts                              
    
\else
    \documentclass{article}

    \usepackage{corl_2024} 
\fi

\usepackage{ifthen}
\usepackage{amsmath}
\usepackage{amsfonts}
\usepackage{graphicx}
\usepackage{bm}
\usepackage{amssymb}
\usepackage{siunitx}
\usepackage{array}
\usepackage{listings}
\usepackage{tabularx}
\usepackage{float}
\usepackage{caption}
\usepackage{tabulary}
\usepackage{booktabs}
\usepackage{amssymb, longtable, array}
\usepackage{tikz}
\usepackage{makecell}
\usepackage{tabularx}
\usepackage{array,multirow}
\usepackage{pgfplots}
\usepackage{subcaption}
\usepackage{placeins}
\usepackage{hyperref}
\hypersetup{
  colorlinks    = true, 
  urlcolor      = blue, 
  linkcolor     = blue, 
  citecolor     = red   
}
\pgfplotsset{
        compat=1.17,
        layers/my layer set/.define layer set={
            background,
            main,
            foreground
        }{
        },
        set layers=my layer set,
    }
\DeclareMathOperator*{\argmin}{argmin} 

\usepackage{tabularray}

\newcommand{\PreserveBackslash}[1]{\let\temp=\\#1\let\\=\temp}

\newcolumntype{C}[1]{>{\PreserveBackslash\centering}p{#1}}
\newcolumntype{R}[1]{>{\PreserveBackslash\raggedleft}p{#1}}
\newcolumntype{L}[1]{>{\PreserveBackslash\raggedright}p{#1}}
\newcolumntype{P}[1]{>{\centering\arraybackslash}p{#1}}
\newcolumntype{M}[1]{>{\centering\arraybackslash}m{#1}}
\usepackage[usestackEOL]{stackengine}

\newcommand{\titlename}{Semantically Safe Robot Manipulation: From Semantic Scene Understanding to Motion Safeguards}
\newcommand{\shorttitle}{Semantically Safe Robot Manipulation}

\newboolean{revised}   
\setboolean{revised}{true}

\newcommand{\edits}{\textcolor{black}}

\ifieee
    \title{\titlename
    }
    
    \author{Lukas Brunke, Yanni Zhang, Ralf R{\"o}mer, Jack Naimer, Nikola Staykov, Siqi Zhou, and Angela P. Schoellig
    \thanks{This work was supported by the \href{https://www.robotics-institute-germany.de/}{Robotics Institute Germany}, funded by BMBF grant 16ME0997K, and by the European Union's Horizon Europe research and innovation programme under the Marie Skłodowska-Curie grant agreement No. 101155035.}
    \thanks{The authors are with the Learning Systems and Robotics Lab and the Munich Institute of Robotics and Machine Intelligence, Technical University of Munich, 80333 Munich, Germany.  
    {Email: \tt\small firstname.lastname@tum.de}}
    \thanks{Lukas Brunke and Angela P. Schoellig are also with the University of Toronto Institute for Aerospace Studies, North York, ON M3H 5T6, Canada, with the University of Toronto Robotics Institute, Toronto, ON M5S 1A4, Canada, and with the Vector Institute for Artificial Intelligence, Toronto, ON M5G 0C6, Canada.}%
    }

\else
    \title{\edits{\titlename}}
    
    \author{
      Jane E.~Doe\\
      Department of Electrical Engineering and Computer Sciences\\
      University of California Berkeley 
      United States\\
      \texttt{janedoe@berkeley.edu} \\
    }
\fi

\begin{document}
\maketitle


\begin{abstract}
Ensuring safe interactions in human-centric environments requires robots to understand and adhere to constraints recognized by humans as ``common sense'' (e.g., ``\textit{moving a cup of water above a laptop is unsafe as the water may spill}" or ``\textit{rotating a cup of water is unsafe as it can lead to pouring its content}"). 
Recent advances in computer vision and machine learning have enabled robots to acquire a semantic understanding of and reason about their operating environments. While extensive literature on safe robot decision-making exists, semantic understanding is rarely integrated into these formulations. In this work, we propose a semantic safety filter framework to certify robot inputs with respect to semantically defined constraints~(e.g., unsafe spatial relationships, behaviors, and poses) and geometrically defined constraints~(e.g., environment-collision and self-collision constraints). In our proposed approach, given perception inputs, we build a semantic map of the 3D environment and leverage the contextual reasoning capabilities of large language models to infer semantically unsafe conditions. These semantically unsafe conditions are then mapped to safe actions through a control barrier certification formulation. \edits{We demonstrate the proposed semantic safety filter in teleoperated manipulation tasks and with learned diffusion policies applied in a real-world kitchen environment that further showcases its effectiveness in addressing practical semantic safety constraints. Together, these experiments highlight our approach's capability to integrate semantics into safety certification, enabling safe robot operation beyond traditional collision avoidance.
}
\end{abstract}

\ifieee
\fi
\renewcommand{\vec}{\bm}
\newcommand{\R}{\mathbb{R}}
\newcommand{\X}{\mathbb{X}}
\newcommand{\U}{\mathbb{U}}

\newcommand{\Csem}{\mathcal{C}_\text{sem}}
\newcommand{\Cenv}{\mathcal{C}_\text{env}}
\newcommand{\Cself}{\mathcal{C}_\text{self}}
\newcommand{\Csemstate}{\mathbb{C}_\text{sem}}
\newcommand{\Cenvstate}{\mathbb{C}_\text{env}}
\newcommand{\Cselfstate}{\mathbb{C}_\text{self}}

\newcommand{\Icam}{\mathbf{I}_{\mathrm{cam}, f}}
\newcommand{\Tcam}{\mathbf{T}_{\mathrm{cam},f}}
\newcommand{\pointcloudfi}{\vec{p}_{f,i}}
\newcommand{\pointcloudi}{\vec{p}_{i}}
\newcommand{\pointcloud}{\vec{p}}
\newcommand{\labelfi}{l_{f,i}}
\newcommand{\labeli}{l_{i}}
\newcommand{\featurefi}{\vec{f}_{f,i}}

\newcommand{\cbfsem}{\vec{h}_\text{sem}}
\newcommand{\cbfsemdot}{\dot{\vec{h}}_\text{sem}}
\newcommand{\cbfenv}{\vec{h}_\text{env}}
\newcommand{\cbfenvdot}{\dot{\vec{h}}_\text{env}}
\newcommand{\cbfself}{\vec{h}_\text{self}}
\newcommand{\cbfselfdot}{\dot{\vec{h}}_\text{self}}

\newcommand{\eepos}{\vec{x}_\text{ee}}
\newcommand{\eevel}{\dot{\vec{x}}_\text{ee}}
\newcommand{\eevelcmd}{\dot{\vec{x}}_\text{ee,cmd}}
\newcommand{\jacobian}{\vec{J}}
\newcommand{\joint}{\vec{q}}
\newcommand{\controlinput}{\vec{u}}
\newcommand{\inputcert}{\vec{u}_\text{cert}}
\newcommand{\inputcmd}{\vec{u}_\text{cmd}}
\newcommand{\state}{\vec{x}}
\newcommand{\forwardkinematics}{\vec{f}_\text{FK}}

\ifthenelse{\boolean{revised}}
{
\ifieee
    \section{INTRODUCTION}
\fi
\label{sec:introduction}
    
    Safety is a key issue in robotics and has been gaining increasing attention across different communities~\cite{brunke2022safe,hsu2023safety}. 
    In safety-critical control, the goal is usually to guarantee set invariance~(i.e., to prevent a system from leaving a certain safe set)~\cite{brunke2022safe}. 
    Based on this definition of safety, various safety filters have been developed in recent years, which can be applied to detect unsafe control inputs and modify them into safe ones in a minimally invasive manner~\cite{hsu2023safety,wabersich2023data}. 
    Existing safety filters such as control barrier function~(CBF) safety filters~\cite{ames2019control} or predictive safety filters~\cite{wabersich2023data} can provide theoretical safety guarantees in terms of set invariance. Still, they assume that the safety constraints are given and explicitly defined in the robot's state space.
    As a result, safety filters in robotics are often restricted to geometrically defined constraints~(e.g., environment-collision constraints). 

    \begin{figure}[t]
    \centering
    \includegraphics[width=\columnwidth]{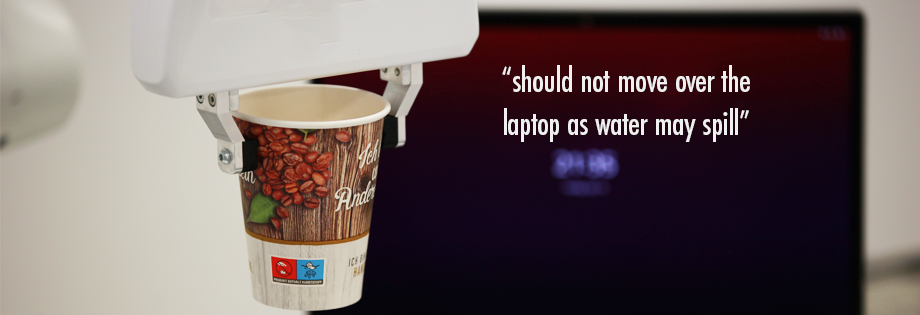}\\[0.1em]
    \includegraphics[width=\columnwidth]{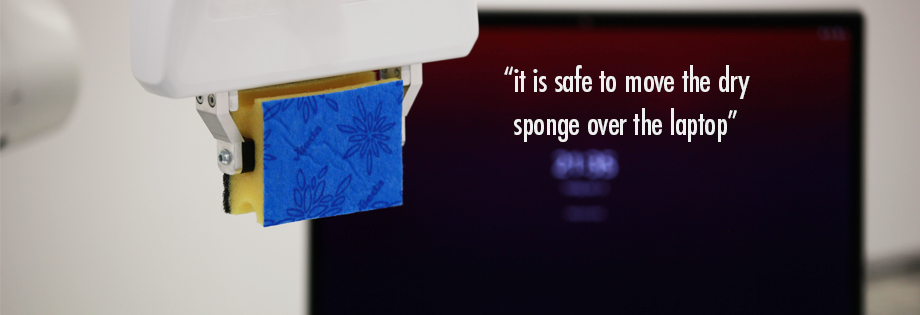}
    \caption{%
    We propose a semantic safety filter framework that leverages semantic scene understanding and contextual reasoning capabilities of large language models to certify robot motions with ``common sense'' constraints. 
    For example, if a manipulator is carrying a cup of water, our proposed semantic safety filter prevents moving the cup above a laptop in the environment to prevent potential spillage~\textit{(top)}. 
    On the contrary, if the robot is tasked to transport a dry sponge, it is allowed to move over a laptop~\textit{(bottom)}. An overview of the work with experiment demonstration results can be found on our website \href{https://utiasdsl.github.io/semantic-manipulation/}{https://utiasdsl.github.io/semantic-manipulation/} and in our short video \href{https://tiny.cc/semantic-manipulation}{https://tiny.cc/semantic-manipulation}.}
    \label{fig:money_figure}
\end{figure}
    
    For robots to operate safely in human-centric environments, they must not only adhere to such geometrically defined constraints but also to constraints that reflect ``common sense''~(see~\autoref{fig:money_figure}).
    In this work, we refer to such constraints as \textit{semantic constraints}. 
    For an example of such semantic constraints, consider a manipulator carrying a filled cup of water over a table.
    To ensure the robot operates safely, it must avoid going over electronic devices due to the risk of spillage. Hence, the semantic constraint should keep the end effector away from the overhead of entities whose semantic labels identify them as electronic devices.
    Additionally, the robot should avoid rotating the cup too much to prevent pouring its content and move slowly close to objects sensitive to water.
    Such semantic constraints are not necessarily ``visible,'' but are critical for real-world applications. 
    Constructing such semantic constraints requires an accurate representation of the 3D environment and a comprehensive understanding of unsafe environment interactions. 

    The development of large language models~(LLMs)~\cite{brown2020language} and vision-language models~(VLMs)~\cite{liu2024visual} has led to significant advances in reasoning about 3D environments~\cite{gu2023conceptgraphs,takmaz2023openmask3d}. \edits{Many recent works leverage these capabilities for language-based decision-making (e.g., to modify robot behavior~\cite{bucker2023latte} or to infer affordability~\cite{yuan2024robopoint}).} However, systematically mapping semantic understanding to constraints remains underexplored. 
    
    In this work, we focus on robot manipulation and present a semantic safety filter that enables robots to reason about and adhere to semantically defined constraints by tightly coupling safe control, 3D perception, and LLMs (see~\autoref{fig:money_figure}).
    Our contributions are as follows:
    \begin{enumerate}
        \item We formulate a semantic CBF safety filter framework that exploits the metric-semantic information from a 3D environment map and reasoning capabilities of LLMs for safe robot manipulation.
        \item Based on environment perception and reasoning, we define three types of semantic constraints:~\textit{(i)} spatial relationship constraints~(e.g., ``do not move the candle below the balloon''),~\textit{(ii)} behavioral constraints~(e.g., ``be slower or more cautious when holding a knife''), and~\textit{(iii)} pose-based constraints~(e.g., ``a cup of water may not be tilted to avoid spillage'').
        \item 
        We demonstrate our framework through hardware experiments using teleoperated and \edits{learned} 
        manipulation tasks. Our results verify the efficacy of our framework in satisfying semantic constraints and highlight the potential of integrating a high-level semantic understanding into safe decision-making.
    \end{enumerate}
        
\ifieee
    \section{RELATED WORK}
\fi
\label{sec:relatedwork}

\subsection{Safe Robot Manipulation} 
In robot control, safety is often defined as ensuring the system does not violate state constraints, which can be achieved by guaranteeing set invariance~\cite{brunke2022safe}.
Traditional approaches achieve safety or collision avoidance through collision-free trajectory generation and high-accuracy tracking control~\cite{spong2022historical}. 
More recently, model predictive control (MPC), learning-based MPC, and geometric control methods have also been applied to enable collision-free manipulation~\cite{chiu2022collision, brunke2022safe,ratliff2018riemannian}. 
Over the past two decades, safety filters, including CBFs~\cite{ames2019control, singletary2022safety}, Hamilton-Jacobi-reachability analysis~\cite{bansal2017hamilton} and predictive control techniques~\cite{wabersich2023data}, have evolved, providing a modular approach to address safe control problems~\cite{brunke2022safe}. Safety filters can be combined with any controllers and certify potentially unsafe control inputs in a minimally invasive manner~\cite{hsu2023safety}.
Existing approaches in safe robot control are often used for geometrically defined constraints~\cite{singletary2022safety,ratliff2018riemannian} and 
often assume the constraints are given ahead of time.
How to translate semantically defined constraints to compatible analytical forms has rarely been addressed in the safe control literature.

\subsection{Semantic 3D Representation and Spatial Reasoning}
Facilitated by advances in machine learning techniques, semantic representations of robots' operating environments can be efficiently distilled from perception inputs~(e.g., through object detection and segmentation)~\cite{ Kirillov_2023_ICCV}. This semantic information has been integrated into 3D mapping and simultaneous localization and mapping~(SLAM) algorithms~\cite{crespo2020semantic} to create consistent instance-level or object-level maps~\cite{Rosinol20icra-Kimera, leutenegger2022okvis2}.
To further facilitate their usage in downstream tasks,  sparse representations such as 3D scene graphs have been proposed as an abstraction of dense metric-semantic maps to capture essential relationships among entities in the environment~\cite{wald2020learning}.
Recent developments in LLMs and VLMs have further enabled open-vocabulary object detection, which has been applied to instance segmentation~\cite{takmaz2023openmask3d} and scene graph generation~\cite{gu2023conceptgraphs}, extending 3D environment representations beyond closed sets of predefined objects. 
\edits{The spatial reasoning capabilities of VLMs~\cite{chen2024spatialvlm, qi2025gpt4scene} have been integrated into 3D mapping frameworks, for example, to identify affordances~\cite{yuan2024robopoint} or relational keypose constraints~\cite{huang2024rekep} for manipulation or to identify safety-critical spatial relationships for navigation~\cite{santos2024updating}.
However, the semantic information in state-of-the-art 3D environment representations has not been fully exploited in downstream safe control tasks.
}

\subsection{Language-Conditioned Robot Decision-Making}
Recently, due to the emergence of foundation models such as 
CLIP~\cite{radford2021learning} and the GPT series~\cite{brown2020language}, there has been a significant advancement in the field of language-conditioned decision-making, including language-aided object grounding~\cite{gu2023conceptgraphs, peng2023openscene}, manipulation~\cite{rashid2023language, xie2023language}\edits{, \cite{oelerich2024language}} and navigation~\cite{huang2023visual}\edits{, \cite{zhang2024tag}}. \edits{The abilities of LLMs and VLMs to understand and output textual information in natural language are used to perform various functions, including} code writing~\cite{huang2023visual, huang2023voxposer}\edits{,\cite{oelerich2024language}}, task planning~\cite{rana2023sayplan, huang2022inner}\edits{,\cite{wang2024apricot}}, verifying robot behavior~\cite{guan2024task}, \edits{and preference learning\cite{wang2024apricot}.}
Hereby, the open-vocabulary capabilities of foundation models are utilized \edits{to enable flexible and adaptive reasoning in unstructured, real-world scenarios. 
Building on these foundations, we leverage LLMs to identify semantically unsafe conditions without being restricted to specific object classes.}

\ifieee
    \section{PROBLEM STATEMENT}
\fi
\label{sec:problem}

In this work, we consider a manipulation setup where objects are arbitrarily placed in the environment, and a robot manipulator is tasked to transport an object in the task space using teleoperation commands or a learned motion policy~(see~\autoref{fig:pipeline}). 
Generally, the teleoperation input or motion policy can be unsafe. Our goal is to design a language-aided safety filter that guarantees safe operation with respect to both semantically defined constraints~$\Csem$~(i.e., spatial relationship-based, behavior-based, and pose-based constraints) and geometrically defined constraints (i.e., environment-collision constraints~$\Cenv$ and self-collision constraints~$\Cself$). We assume that the system can perceive and reason about its environment through a set of RGB-D images~$\{\Icam\}$ of the scene and the associated camera poses~$\{\Tcam\}$, where~$f$ denotes the frame index.

\begin{figure*}[ht]
\begin{center}
  \includegraphics[width=\textwidth]{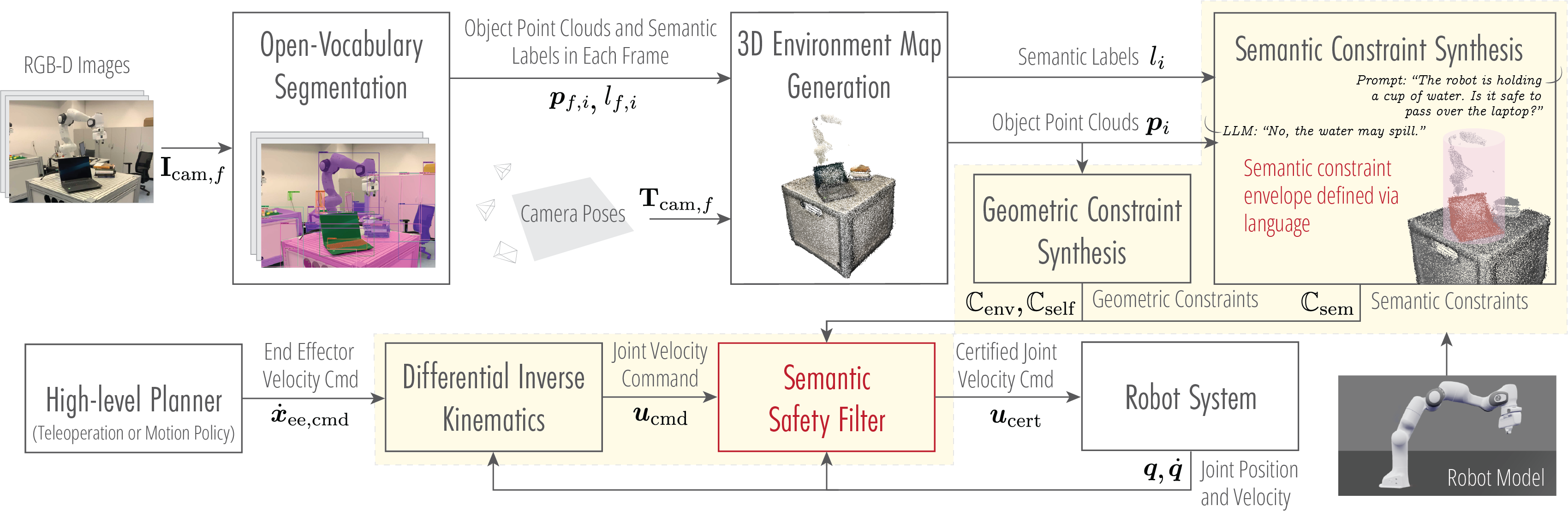}
    \caption{An overview of our proposed semantic safety filter framework. The perception module segments the visual input and builds a semantic world representation. The LLM is queried based on the list of semantic labels and the manipulated object. It outputs the semantic context~$\mathcal{S}$, which contains a list of unsafe spatial relationship-based semantic constraints for each object in the scene, a list of behavioral-based semantic constraints, and a pose-based semantic constraint. The semantic context, together with the point clouds of the objects in the scene, are then used to define safe sets for our proposed semantic safety filter. Additionally, based on the semantic context, the safety filter's parameters are adapted, for example, to prevent end effector rotations or to approach certain objects more carefully. At each time step, 
    a high-level uncertified command from a human operator or a motion policy is mapped to the joint velocity $\inputcmd$ through differential inverse kinematics, certified by the proposed semantic safety filter, and then sent to the robot system.
    }
  \label{fig:pipeline} 
\end{center}
\end{figure*}

 We note that the term semantic constraint has scenario-dependent definitions in the literature~(e.g., grasp types and trajectory constraints for robotic hands~\cite{li2020transferring}). We refer to semantic constraints as the task-space constraints on a robotic manipulator's end effector that are related to high-level semantic concepts (e.g., \textit{``not moving a filled cup of water over electronic devices''} and \textit{``not rotating a cup of water to avoid spilling its content''}). In contrast to typical collision avoidance constraints, semantically unsafe states are not necessarily ``visible'' (i.e., occupied by objects), and synthesizing the semantic constraints requires a high-level understanding of the environment and the manipulated object. In this work, we leverage the perception inputs, a model of the robot system, and an LLM to design a safety filter that guarantees semantic safety while also avoiding self-collisions and collisions with the environment.

\ifieee
    \section{METHODOLOGY}
\fi
In this section, we present the components of our proposed semantic safety filter framework, which is visualized in~\autoref{fig:pipeline}. Given a set of RGB-D images and the associated camera poses, we first generate a semantic map of the 3D environment (\autoref{subsec:3d_environment_map}). Then, a set of semantic constraints is synthesized using the semantic map and the LLM (\autoref{subsec:semantic_constraints}). Finally, a semantic safety filter is formulated to account for the semantic 
constraints~(\autoref{subsec:safety_filter}).

\subsection{3D Environment Map Generation}
\label{subsec:3d_environment_map}
The semantic constraints synthesis depends on a 3D environment representation that supports semantic reasoning for downstream planning and control tasks. This motivates a language-embedded representation approach. In this work, we 
construct an open-vocabulary object-level representation of the 3D environment~\cite{gu2023conceptgraphs,takmaz2023openmask3d} to aid our safety filter design. 

The input to the 3D environment map generation module is a set of RGB-D frames~$\{\Icam\}$ along with the camera poses~$\{\Tcam\}$. \edits{The RGB-D images are segmented~\cite{Kirillov_2023_ICCV}, and every resulting segmentation mask is embedded through the CLIP visual encoder~\cite{radford2021learning} to generate segmented point clouds $\pointcloudfi$ and their associated visual embeddings~$\vec{f}_{f,i}$ for each object~$i$ in each frame~$f$.} The segmented object-level point clouds~$\pointcloudfi$ together with the associated camera poses~$\Tcam$ and feature vectors~$\featurefi$ are then used to associate objects across multiple views based on geometric and semantic similarities~\cite{gu2023conceptgraphs}. The per-frame information is incrementally fused to create a consistent object-level point-cloud representation of the 3D environment. The output of the map is a set of point clouds~$\pointcloudi$ and embeddings~$\vec{f}_i$ for each object in the scene. \edits{Similar to~\cite{takmaz2023openmask3d,gu2023conceptgraphs}, we assign labels~$l_i$ to objects by comparing the cosine similarity between the object embeddings~$\vec{f}_i$ and the text embeddings derived from the list of object categories in the ScanNet200 dataset~\cite{rozenberszki2022language}. The object’s class is assigned based on the pair of embeddings with the highest cosine similarity score.} 

\subsection{Semantic Constraint Synthesis}
\label{subsec:semantic_constraints}

We distinguish among three types of semantic safety: \textit{(i)} unsafe spatial relationships between the object manipulated by the robot and the objects in the scene~(e.g., ``do not move the candle below the balloon"), \textit{(ii)} behavioral constraints, such as constraints on the end effector velocity based on the manipulated object and the scene objects~(e.g., ``be slower or more cautious when holding a knife''), and \textit{(iii)} pose constraints on the end effector dependent on the manipulated object~(e.g., ``keep the cup of water upright to avoid spillage'').
Such semantic constraints are object- and scene-dependent and tedious to specify manually. Therefore, we employ an LLM to synthesize them in an automated manner.

\edits{We design a language prompt for the LLM, which consists of multiple in-context examples and a final request as the true query.} \edits{For each object in the scene, the requests contain the following components: \textit{(i)} a high-level description of the scene~\edits{specified by the user directly} (or, inferred from a small set of images via VLM), \textit{(ii)} the object the robot is manipulating, and \textit{(iii)} the object itself.} 
Using these requests, we determine three sets of semantic constraints. 
First, the set of unsafe spatial relationships is $\mathcal{S}_{\text{r}}(o) = \left\{(l_i, r_i)\right\}_{i=1}^{N_{\text{r}}}$, where $o$ is the manipulated object~(e.g., \texttt{cup of water}), $l_i$ is an object in the scene~(e.g., \texttt{laptop}, \texttt{book}, etc.), $r_i$ is an unsafe spatial relationship~(e.g., \texttt{above}, \texttt{under}, or \texttt{around}), and~$N_\text{r}$ is the number of unsafe spatial relationships. 
Second, the set of unsafe behaviors is $\mathcal{S}_{\text{b}}(o) = \left\{(l_i, b_i)\right\}_{i=1}^{N_{\text{b}}}$, where $b_i$ indicates \texttt{caution} or \texttt{no caution} and $N_{\text{b}}$ is the number of unsafe behaviors. 
Finally, the pose-based constraint set is $\mathcal{S}_{\text{T}}(o) = \left\{T\right\}$, where $T$ specifies the end effector orientation constraint~(\texttt{\edits{constrained} rotation} or \texttt{free rotation}). 
The set of semantic constraints is the union of all the semantic constraints listed above: $\mathcal{S}(o) = \mathcal{S}_{\text{r}}(o) \cup \mathcal{S}_{\text{b}}(o) \cup \mathcal{S}_{\text{T}}(o)$. 
For the $o = \texttt{cup of water}$ transportation example in the scene with only $l_0 = \texttt{laptop}$, we have $\mathcal{S}_{\text{r}}(o) = \{(\texttt{laptop}, \texttt{above})\}$, $\mathcal{S}_{\text{b}}(o) = \{(\texttt{laptop}, \texttt{caution}) \} $, and $\mathcal{S}_{\text{T}}(o) = \{ \texttt{\edits{constrained} rotation} \}$. 

Our proposed semantic safety filter is designed based on control barrier certification~\cite{ames2019control}. 
In the following, we describe how we design the CBF safety filter using~$\mathcal{S}(o)$. 
We denote the joint positions by~$\vec{q}\in\R^n$~(with $n = 7$ in our case) and, similar to~\cite{singletary2022safety}, assume direct control over the joint velocity~$\dot{\vec{q}}$,~(i.e., $\dot{\vec{q}} = \vec{u}$), which can be achieved via standard lower-level motion control techniques~\cite{lynch2017modern}. The robot's end effector position and velocity can be related to its joint position and velocity as
$\eepos = \forwardkinematics(\vec{q})$ and $\eevel = \vec{J}(\vec{q})\: \dot{\vec{q}}$,
where $\forwardkinematics:\R^n\mapsto \R^3$ and ${\vec{J}(\vec{q}) \in \mathbb{R}^{3 \times n}}$ are the translational component of the forward kinematics and the associated Jacobian matrix, respectively.

\subsubsection{Spatial Relationship Constraints}
The semantic constraint sets are parameterized as the zero superlevel sets of continuously differentiable functions~$\cbfsem$. Intuitively, the CBF certification framework ensures the positive invariance of the semantically safe set. This means that if the robot does not violate the semantic constraint initially, it will not violate it for all future times. 
For each pair $(l_i, r_i)$ in $\mathcal{S}_{\text{r}}(o)$, based on the point cloud
$\vec{p}_{i}$ of the object $l_i$ and the undesirable spatial relationship~$r_i$, we define a differentiable function~$g_i:\R^3 \mapsto \R$ to capture the set of points
which the robot end effector should not move into to preserve semantic safety. The semantically safe set can be expressed as
\begin{align*}
    \Csemstate = 
    \left\{\eepos \in \mathbb{R}^3 \left| \:
    g_i(\eepos; \vec \theta_i) \geq 1, \; i = 1,\dots,N_\text{r} \right. \right\},
\end{align*}
where $\eepos=[x,y,z]^\mathsf{T}\in\R^3$ denotes the end effector position and $\vec \theta_i$ are parameters dependent on the object point cloud $\pointcloud_i$ and the relationship $r_i$. 

\begin{figure}[t]
    \centering
    \includegraphics[width=\columnwidth]{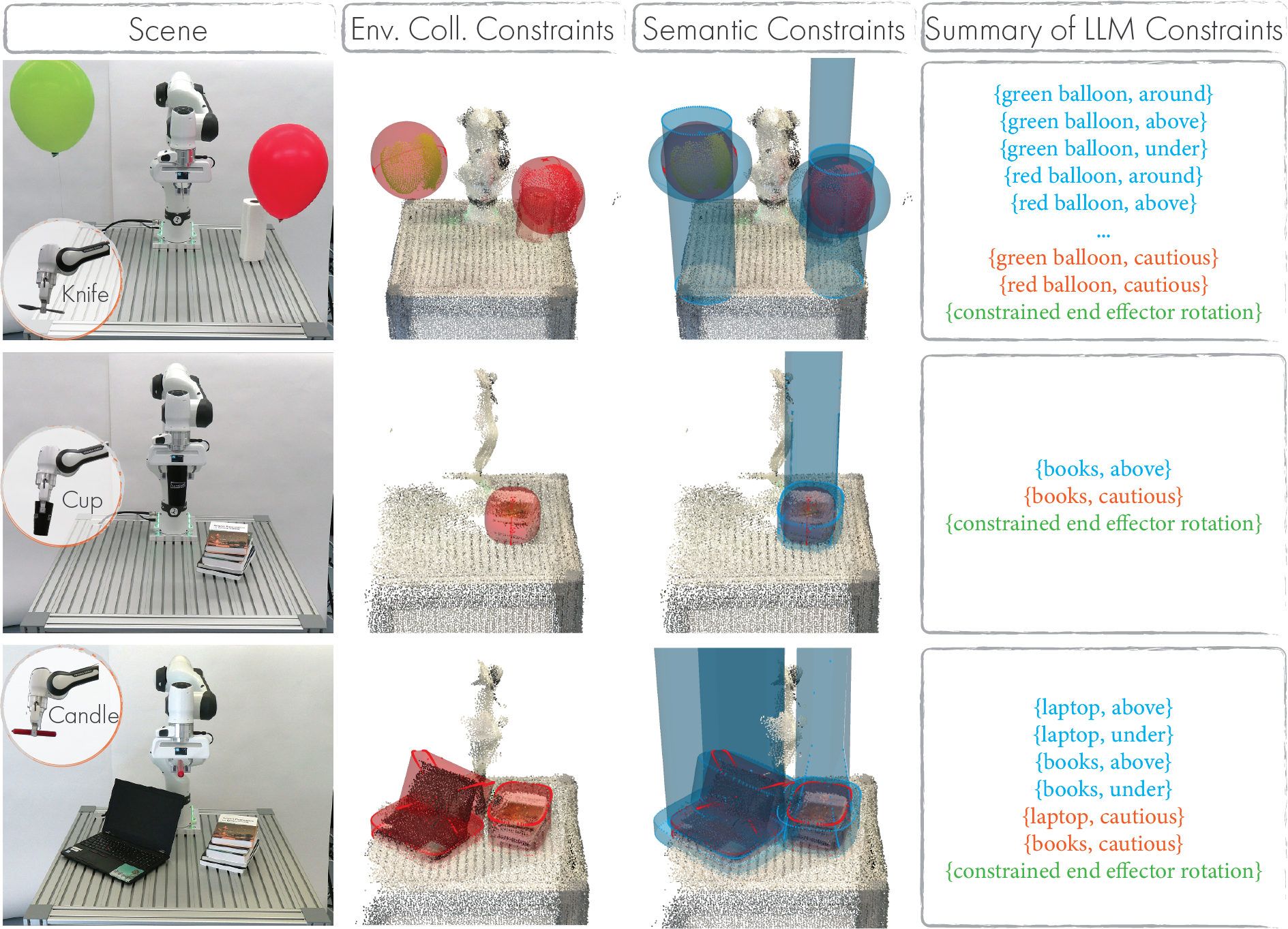}
    \caption{Examples of the environment collision and semantic constraints enforced by our proposed semantic safety filter. For each scene, environment collision constraints are generated based on the point clouds of individual objects while the semantic constraints are synthesized based on the point clouds and labels of individual objects as well as the semantic safety conditions from the LLM. The semantic safety conditions are further categorized into spatial relationship constraints~(blue text), behavioral constraints~(orange text), and end effector pose constraints~(green text).}
    \label{fig:llm_reasoning}
\end{figure}

For the $\{\texttt{laptop},\texttt{above}\}$ example (as also illustrated in~\autoref{fig:llm_reasoning}), we define the semantically unsafe sets as a differentiable approximation 
using a superquadric~\cite{liu2022robust}:
\begin{align*}
    &g_i(\eepos; \vec{\theta_i}) =
    \\
    &\left( 
        \left( \frac{\tau_1(\eepos)}{a_{x,i}} \right)^{\frac{2}{\epsilon_{2,i}}} + 
        \left( \frac{\tau_2(\eepos)}{a_{y,i}} \right)^{\frac{2}{\epsilon_{2,i}}} 
    \right)^{\frac{\epsilon_{2,i}}{\epsilon_{1,i}}} \hspace{-0.2cm}+ 
    \left( 
        \frac{\tau_3(\eepos)}{a_{z,i}} 
    \right)^{\frac{2}{\epsilon_{1,i}}} \,,
\end{align*}
where $\epsilon_{1,i}$ and $\epsilon_{2,i}$ define the shape of the superquadric, $a_{x,i}, a_{y,i},$ and $a_{z,i}$ are scaling parameters, 
and $\tau_1$, $\tau_2$, and $\tau_3$ transform the end effector coordinates into the superquadric's coordinate frame.
To improve nonconvex objects' representations, we create unions of superquadrics to accurately fit spatial constraints. For example, we fit separate superquadrics for the part of the \texttt{laptop}'s point cloud that resembles the keyboard and the screen. This segmentation by parts can be achieved using plane detection algorithms or learned segmentation models~\cite{sharma2022prifit}. 
To account for the spatial relationship \texttt{above}, we extend the point cloud in its positive $z$-direction. For this, we duplicate the point cloud, set the duplicate's $z$-coordinates to be outside the robot's workspace, and fit the superquadric based on the union of the original and the expanded point cloud.
\edits{We consider 12 spatial relationships in total,} such as $\texttt{under}$ and $\texttt{around}$, for which we define similar superquadrics. 

To achieve spatial semantic safety with respect to the semantic constraint set~$\mathbb{C}_\text{sem}$, we define a vector of CBFs
$\vec{h}_\text{sem}(\eepos)$, where the $i$-th element is
\begin{align}
    \label{eq:semantic_cbf}
    h_{\text{sem},i}(\eepos) = g_i(\eepos;\vec \theta_i) - 1.
\end{align}
Using forward kinematics, we can express the semantic constraint set based on the CBFs~\eqref{eq:semantic_cbf} in the robot's configuration space, which yields our desired safe set
\begin{equation}
    \mathbb{C}_\text{sem} = \{ \vec{q} \in \mathbb{R}^n \left| \: \vec{h}_\text{sem}(\forwardkinematics(\vec{q})) \geq \vec{0} \right. \}\,.
    \label{eqn:safe_set}
\end{equation}

\subsubsection{Behavioral Constraints}
The behavioral constraints are implemented using constraints on the time derivative of the CBF, i.e., the control invariance condition~\cite{ames2019control}, of the form
\begin{align*}
    \cbfsemdot(\vec{q},\controlinput) = \mathbf{H}_{\text{sem}} (\vec{q}) \jacobian(\vec{q}) \:\controlinput\ge -\boldsymbol{\alpha}_\text{sem} (\cbfsem(\vec{q}); \mathcal{S}_{\text{b}}(o)),
\end{align*}
where $\mathbf{H}_{\text{sem}}(\vec{q}) = \left .\frac{\partial \cbfsem }{\partial\eepos}\right \vert_{\eepos=\forwardkinematics(\vec{q})}$ and $\boldsymbol{\alpha}_\text{sem}$ is a vector of class $\mathcal{K}_\infty$ functions~(i.e., real-valued functions that pass through the origin and are strictly increasing). Intuitively, the condition bounds how fast the robot system is allowed to approach the semantic safety boundary through the design of $\boldsymbol{\alpha}_\text{sem}$ and ensures that the constraints defined by $\cbfsem$ are always satisfied (i.e., the set~$\Csem$ is forward invariant)~\cite{ames2019control}. In particular, we design the class $\mathcal{K}_\infty$ to adhere to behavioral semantic constraints~$b_j$ from $\mathcal{S}_{\text{b}}(o)$ such that the system approaches the safe set boundary of the object with label $l_j$ more slowly and exhibits the desired level of caution. 
For example, for the case $b_j = \texttt{caution}$, we reduce the steepness of $\alpha_{\text{sem},j}$. 
In that case, we also write $\alpha_{\text{sem},j}(\cdot; \texttt{caution}) = \alpha_{\text{sem},\text{c},j}(\cdot)$. 
This reduction can be achieved by using a class $\mathcal{K}_\infty$ that is strictly smaller than $\alpha_{\text{sem},j}$ for positive $h_{\text{sem},j}$. Such a function can be produced by multiplying the function $\alpha_{\text{sem},j}$ with a scalar $w_{\alpha, j} \in \left(0, 1\right)$. 

\subsubsection{Pose Constraints}
The pose constraint is active if $\mathcal{S}_{\text{T}}(o) = \left\{ \texttt{\edits{constrained} rotation} \right\}$. In that case, we add the following constraint:
\begin{equation*}
    \vec \Delta \vec\psi_{\text{min}} \leq \log(\vec R_\text{des}\vec R_\text{cur}^\mathsf{T})^{\vee} - \vec \psi \leq \vec \Delta \vec \psi_{\text{max}}\,,
\end{equation*}
where $\vec R_\text{des}$ is the desired rotation of the end effector~(the end effector's initial orientation during the object's pick-up), $\vec R_\text{cur}$ is the current rotation of the end effector, $\vec{\psi} = \vec{J}_o(\vec{q}) \vec u \Delta t$ is the predicted rotation of the end effector at the next timestep~$(t + \Delta t)$ with $\vec J_o(\vec{q})$ being the Jacobian relating the joint velocity to the angular velocity of the end effector, $(\cdot)^{\vee}$ denotes the inverse of the skew-symmetric operator $(\cdot)^{\wedge}$~\cite{barfoot2024state}, and $\vec \Delta \vec \psi_{\text{min}}$ and $\vec \Delta \vec \psi_{\text{max}}$ are the tolerated orientation errors.
In our implementation, we leverage a softened formulation for this constraint to make the approach less prone to infeasibility. 
We express the softened pose constraint using the objective
    $\vec{w}_\text{rot}(\mathcal{S}_{\text{T}}(o))^\mathsf{T} \vec{L}_\text{rot}(\vec{q}, \vec{u})$.
The weight $\vec{w}_{\text{rot}}\in\R^2$ is determined based on the semantic context $T$ in $\mathcal{S}_{\text{T}}$. The end effector is free to rotate if $T = \texttt{free rotation}$~(e.g., no object is being held) with $\vec{w}_{\text{rot}} = 0$, but $\vec{w}_{\text{rot}} > 0$ if $T = \texttt{\edits{constrained} rotation}$~(e.g., a cup of water is being manipulated to prevent spilling). The vector $\vec{L}_\text{rot}$ is
\begin{align*}
    \vec{L}_\text{rot}(\vec{q}, \vec{u}) 
    = \begin{bmatrix}
        \Vert \log(\vec R_\text{des}\vec R_\text{cur}^\mathsf{T})^{\vee} - \vec \psi \Vert _2^2 & 
        \Vert \vec \psi \Vert _2^2
    \end{bmatrix}^\mathsf{T} \,,
\end{align*}
where the first element represents the cost for the difference between the predicted orientation at the next timestep and the desired orientation of the manipulator's end effector and
the purpose of the second element is to prevent the end effector from rotating too fast and to keep perturbations small. 

\subsection{Geometric Constraints}
In addition to semantic constraints, we require the robot to adhere to geometric constraints, which include environment-collision and self-collision constraints. 
We incorporate these additional constraints into two more vectors of CBFs ${\vec{h}_\text{env}(\vec q)
}$ and~$\vec{h}_\text{self} (\vec q)
$. The environment-collision constraints are defined based on CBFs using superquadrics fitted to the point clouds $\pointcloudi$~(see previous section); the self-collision constraints are formulated by placing multiple spherical CBFs along the body of the robot, similarly as in~\cite{singletary2022safety}.

\subsection{Semantic Safety Filter Formulation}
\label{subsec:safety_filter}
Given the semantic constraints~$\Csem$ and the set $\mathcal{S}$, our goal is to modify potentially unsafe commands sent by a human operator or coming from a motion policy. As depicted in~\autoref{fig:pipeline}, in our setup, we send the desired end effector velocity commands~$\eevelcmd$, which are converted to desired joint velocity commands~$\inputcmd$ using differential inverse kinematics.
The semantic safety filter then computes a certified input~$\inputcert$ that best matches the desired joint velocity~$\inputcmd$ while ensuring semantic and geometric constraint satisfaction. 
The semantic safety filter is formulated as
\begin{equation}
    \begin{split}
    \raisetag{30pt}
    \label{eqn:cbf_qp}
        \vec{u}_\text{cert} = \argmin_{\vec{u} \in \U} \quad \Vert \vec{u} - \vec{u}_\text{cmd} \Vert^2_2 &+ \vec{w}_\text{rot}(\mathcal{S}_{\text{T}}(o))^\mathsf{T} \vec{L}_\text{rot}(\vec{q}, \vec{u})
  \\
        \text{s.\,t.} \quad \vec{\dot{h}}_\text{sem}(\vec{q},\controlinput; \mathcal{S}_{\text{r}}(o))
        &\geq - \vec{\alpha}_\text{sem}(\vec{h}_\text{sem}(\vec{q}); \mathcal{S}_{\text{b}}(o))\\
        \quad \vec{\dot{h}}_\text{env}(\vec{q},\controlinput) 
        &\geq - \vec{\alpha}_\text{env}(\vec{h}_\text{env}(\vec{q}); \mathcal{S}_{\text{b}}(o)) \\
        \quad \vec{\dot{h}}_\text{self}(\vec{q},\controlinput) 
        &\geq - \vec{\alpha}_\text{self}(\vec{h}_\text{self}(\vec{q})) \\
        \quad \vec{\dot{h}}_\text{lim}(\vec{q},\controlinput) &\geq - \vec{\alpha}_\text{lim}(\vec{h}_\text{lim}(\vec{q})) \,,
    \end{split}
\end{equation}
where we made the dependency on the semantic context~$\mathcal{S}(o)$ explicit, added joint angle and velocity constraints through additional CBFs~$\vec{h}_\text{lim}(\vec{q})$, and $\vec{\alpha}_\text{env}$, $\vec{\alpha}_\text{self}$, $\vec{\alpha}_\text{lim} \in \mathcal{K}_\infty$.
The first term in the cost function minimizes the difference between the certified input and the desired input command, while the second term penalizes rotations away from the desired rotation. 
The four sets of inequality constraints in~\eqref{eqn:cbf_qp} correspond to the semantic spatial relationship-based, environment-collision, self-collision, and joint angle and velocity constraints. The class $\mathcal{K}_\infty$ functions define behavioral semantics for each constraint, and the objective provides softened posed-based safety constraints. 
The semantic safety filter optimization problem~\eqref{eqn:cbf_qp} is a QP that can be efficiently solved online.
Overall, the semantic safety filter in~\eqref{eqn:cbf_qp} finds the control input that best matches the desired input while ensuring all constraints are satisfied.

\ifieee
    \section{EXPERIMENTS}
\fi
In this section, we present the experimental evaluation of our proposed semantic safety filter. In the real-world experiment, a Franka Emika FR3 robotic manipulator is deployed with our proposed semantic safety filter in a closed loop to prevent potentially unsafe commands from a non-expert user or a learned motion policy. A video of the experimental results can be found at \href{https://tiny.cc/semantic-manipulation}{https://tiny.cc/semantic-manipulation} and on our website \href{https://utiasdsl.github.io/semantic-manipulation/}{https://utiasdsl.github.io/semantic-manipulation/}.

\subsection{Semantic Perception}
\label{sec:experiment_perception}
In our evaluation, we consider static~(unless manipulated by the robot) scenes, some of which are visualized in~\autoref{fig:llm_reasoning} and various manipulated objects, including a \texttt{dry sponge}, a \texttt{cup of water}, a \texttt{lit candle}, and a \texttt{knife}. The geometries of the manipulated objects and the robot are assumed to be known. However, the environment in which the robot operates is assumed to be unknown; 
a map for each environment is generated using RGB-D images and associated camera frames as described in \autoref{subsec:3d_environment_map}.  The RGB-D images were recorded using a Femto Bolt
and the camera poses were obtained by running visual-inertial SLAM~\cite{Seiskari_2022_WACV}. Each scene was reconstructed using approximately 50 to 200 RGB-D images \edits{and their associated camera poses}, and the semantics are determined as described in~\autoref{subsec:3d_environment_map}. 
Examples of the reconstructed scenes are shown in~\autoref{fig:llm_reasoning}.

\subsection{LLM Prompting}
\label{sec:llm_prompting}
\begin{table}[t]
\caption{\edits{The multi-prompt strategy yields higher precision and recall than single-prompting on our benchmarking dataset of ground-truth constraints.}}
    \centering
\label{tab:llm-prompting}
\begin{tabular}{lcc}
\hline\hline
\textbf{Prompting Strategy}          & \textbf{Precision (\%)} & \textbf{Recall (\%)} \\ \hline
Single-Prompt         & 29                                 & 78    \\
Multi-Prompt      & 60                                 & 99                        \\\hline\hline\\[-0.5em]
\end{tabular}
\end{table}

We created a benchmarking dataset of objects, scenes, and ground-truth constraints to evaluate the semantic constraint generation. The dataset includes over 50 semantic constraints containing all semantic constraint types, as well as objects and scenes not encountered in our experiments, and we use it to evaluate two different prompting strategies on an LLM~(GPT-4o~\cite{achiam2023gpt}). The first strategy (single-prompt) requests the full set $\mathcal{S}(o)$ at once, while the second strategy (multi-prompt) requests only one pair or a singleton~(for the semantic pose constraint) for each prompt.  
The multi-prompt method proved more accurate, as indicated by the higher precision and recall in~\autoref{tab:llm-prompting}.
We adjusted the final prompt until the desired level of accuracy was achieved on the validation dataset. 

For our robot experiments, we follow the methodologies in \autoref{subsec:semantic_constraints} to identify semantically unsafe object-relationship pairs, behaviors, and poses. Examples are shown in the last column of~\autoref{fig:llm_reasoning}. We query the LLM for each object-relationship pair for each scene multiple times using majority voting to determine if the spatial relationship between the manipulated object and the particular object in the scene is semantically safe. We run additional queries to determine if the object held by the manipulator may be rotated and if increased caution should be exhibited close to each of the objects in the scene. These responses are then used in combination with each object's point cloud to determine the constraint envelopes~(see~\autoref{fig:llm_reasoning}), the class $\mathcal{K}_\infty$ function, and the weight $\vec{w}_\text{rot}$. 

\subsection{\edits{Demonstration in Tabletop Manipulation Tasks}}
Using our semantic safety filter, we execute various teleoperation and pick-and-place tasks on the robot. 
We run our semantic safety filter at $\SI{45}{\hertz}$.
\begin{table*}[t]
\centering
    \caption{%
    A summary table of the mean percentages and their associated standard deviations of time steps that violate any of the constraints $\mathcal{C}_{\text{sem}}$, $\mathcal{C}_{\text{env}}, \mathcal{C}_{\text{self}}, \mathcal{C}_{\text{lim}}$. Our evaluation includes a baseline without a safety filter, a safety filter accounting for geometric constraints, and our proposed semantic safety filter. We use three scenes and five different manipulation cases~(four objects and empty-handed) with five teleoperated trajectories each, resulting in a total of 40 trajectories for each method. 
    Each combination of objects and scenes yielded different geometric and semantic constraints. 
    }
    \label{tab:additional_eval}
    \begin{tabular}{ l c c c c c}
        \toprule
        Scene & Held Object$^\dagger$ & No Safety Filter  & Nominal Safety Filter (w/o $\mathcal{C}_{\text{sem}}$) & \textbf{Our Semantic Safety Filter} \\

        \midrule
        
        \multirow{2}{*}{\texttt{\textcolor{black}{\{books\}}}} & \texttt{dry sponge}  &   11.06\% $\pm$ 13.60\% &  \textbf{\phantom{0}0.00\% $\pm$ \phantom{0}0.00\%} &  \textbf{\phantom{0}0.00\% $\pm$ \phantom{0}0.00\%} &
        \\\cline{2-6}
       & \texttt{\textcolor{red}{cup of water}} & 70.37\% $\pm$ 23.51\% &  64.98\% $\pm$ 33.42\%  & 
       \textbf{\phantom{0}0.00\% $\pm$ \phantom{0}0.00\%} & \\
        \hline

        \multirow{3}{*}{\texttt{\textcolor{black}{\{laptop, books\}}}} & \texttt{none}  &   36.29\% $\pm$ 18.29\% &  \textbf{\phantom{0}0.00\% $\pm$ \phantom{0}0.00\%} & \textbf{\phantom{0}0.00\% $\pm$ \phantom{0}0.00\%} &
        \\\cline{2-6}
       & \texttt{\textcolor{red}{lit candle}} & 65.21\% $\pm$ 14.20\% &  51.33\% $\pm$ 27.85\%  & 
       \textbf{\phantom{0}0.00\% $\pm$ \phantom{0}0.00\%} & 
        \\\cline{2-6}
       & \texttt{\textcolor{red}{cup of water}} & 59.40\% $\pm$ 12.02\% &  41.90\% $\pm$ 25.46\%  & 
       \textbf{\phantom{0}0.00\% $\pm$ \phantom{0}0.00\%} & \\
        \hline

        \multirow{3}{*}{\texttt{\textcolor{black}{\{balloons, paper towel\}}}} & \texttt{cup of water}  &   28.07\% $\pm$ 14.77\% &  \textbf{\phantom{0}0.00\% $\pm$ \phantom{0}0.00\%} &  \textbf{\phantom{0}0.00\% $\pm$ \phantom{0}0.00\%} &
        \\\cline{2-6}
       & \texttt{\textcolor{red}{lit candle}} & 50.33\% $\pm$ \phantom{0}9.44\% &  49.89\% $\pm$ \phantom{0}9.04\%  & 
       \textbf{\phantom{0}0.00\% $\pm$ \phantom{0}0.00\%} & 
        \\\cline{2-6}
       & \texttt{\textcolor{red}{knife}} & 49.07\% $\pm$ 16.16\% &   30.85\% $\pm$ 10.53\%  & 
       \textbf{\phantom{0}0.00\% $\pm$ \phantom{0}0.00\%} & \\
        \bottomrule
    \end{tabular}
    \flushleft{\footnotesize $^\dagger$The objects in \textcolor{red}{red} result in semantic constraints.}
\end{table*}
Our teleoperation experiments are summarized in~\autoref{tab:additional_eval}. The teleoperation commands are provided through a teleoperation interface as end effector velocities in the Cartesian space and smoothed using a low-pass filter. We calculate the associated joint velocities with differential inverse kinematics. Each scene is tested with multiple held objects, which require different sets of semantic constraints~(see~\autoref{fig:llm_reasoning}). The results in the table confirm that our safety filters can effectively account for collision avoidance constraints and any semantic constraints generated by our synthesis module, as no constraint violations occur in any of our experiments when the safety filter is active. 
\begin{figure}
    \centering
    \input{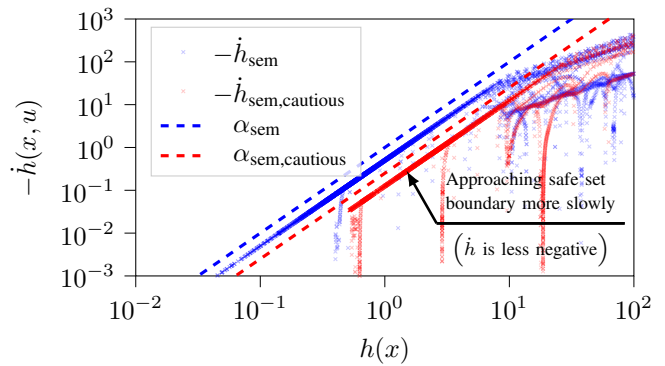}
    \caption{
    The level of caution determines how quickly the end effector approaches a safety constraint boundary. In the \texttt{books} scene, we increase caution by adjusting the class $\mathcal{K}_\infty$ function when holding a \texttt{cup of water} under the same semantic constraint during teleoperation. In the cautious case, the negative time derivatives remain below the red dashed line, satisfying the CBF condition. Since  $\alpha_{\text{sem,c}} < \alpha_{\text{sem}}$, the end effector approaches the boundary more slowly. 
    Note that the $y$-axis is inverted.}
    \label{fig:caution}
\end{figure}

We highlight how the different levels of caution determine how quickly the end effector holding a specific object may approach the boundary of a safety constraint boundary. For the scene \texttt{books}, we show increased caution by modifying the class $\mathcal{K}_\infty$ function when holding a \texttt{cup of water} for the same semantic constraint during teleoperation.
For the cautious case, the negative time derivatives~$(-\dot{h})$~(red) stay below the red dashed line, confirming the CBF condition's satisfaction. As $\alpha_{\text{sem,c}}(h) = \frac{1}{4}h^2$ is strictly smaller than $\alpha_{\text{sem}}(h) = h^2$ on $h > 0$, the end effector approaches the boundary of this semantic constraint slower. 
Note that we manually overwrote the level of caution for this particular demonstration to compare the closed-loop behavior on the same semantic CBF constraint. Generally, the level of caution is determined through the method outlined in~\autoref{subsec:semantic_constraints}.
\begin{figure}
    \centering
    \input{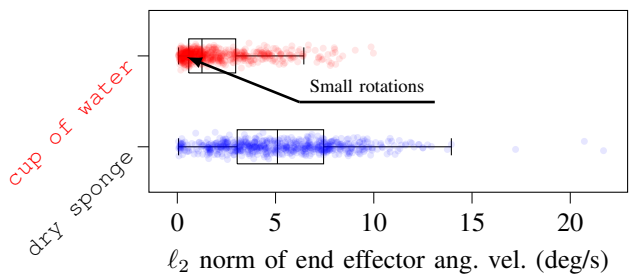}
    \caption{Demonstration of the active~(inactive) rotation \edits{constraint} when the robot is holding a \texttt{cup of water}~(\texttt{dry sponge}) in the scene \texttt{books}. The distribution for the \texttt{cup of water} is skewed towards smaller angular velocities; an active rotation \edits{constraint} (red) generally yields reduced end effector rotations as compared to the inactive case (blue). 
     }
    \label{fig:ang-vel}
\end{figure}

Finally, we demonstrate the effectiveness of \edits{constraining} rotations for different objects based on their semantics in~\autoref{fig:ang-vel}. Our semantic safety filter successfully reduces the median of the norm of the end effector's angular velocity by $75.39\%$ if the rotation \edits{constraint} is active~(see \texttt{cup of water}). The box plot also highlights that the interquartile range of the end effector's angular velocity norm is reduced by $45.67\%$ compared to the robot holding the \texttt{dry sponge}. \edits{We note that, in our implementation, the semantic context $\mathcal{S}_T(o)$ is binary (i.e., either constrained or unconstrained rotation). However, it is generally possible to prompt the LLM with finer granularity and enforce varying levels of cautiousness by appropriately configuring the vector $\boldsymbol{\omega}_\text{rot}$.}

\edits{To further evaluate the scalability of our proposed approach to more complex environments, we applied our semantic safety filter to pick-and-place tasks in a cluttered environment with~17 objects~(see our supplementary video). 
When the robot is holding the \texttt{dry sponge}, its end effector is allowed to rotate and move the object above electronic devices with no additional caution considered; in contrast, the robot's motion is much more constrained when holding the \texttt{cup of water} to prevent potential spillage. 
} 

\subsection{\edits{Demonstration in a Real-World Kitchen Environment}}

\edits{To demonstrate the applicability of our proposed filter beyond teleoperation, we conducted experiments in a real-world kitchen environment and trained diffusion policies~\cite{Chi-RSS-23} for five different transportation tasks involving various semantically unsafe constraints. These constraints include handling fragile items and preventing fire and electrical hazards. Clips of this set of experiments are included in the supplementary video. \autoref{fig:kitchen_results} compares the normalized CBFs for our proposed semantic safety filter and a nominal geometric safety filter that does not account for semantic constraints. The proposed semantic safety filter successfully prevents unsafe actions such as placing a metal cup inside a microwave or putting a pressurized spray can on a stove. This set of experiments highlights the generalizability of our proposed approach to learned policies and its applicability in real-world settings.}

\begin{figure}
    \centering
    \includegraphics[width=\columnwidth]{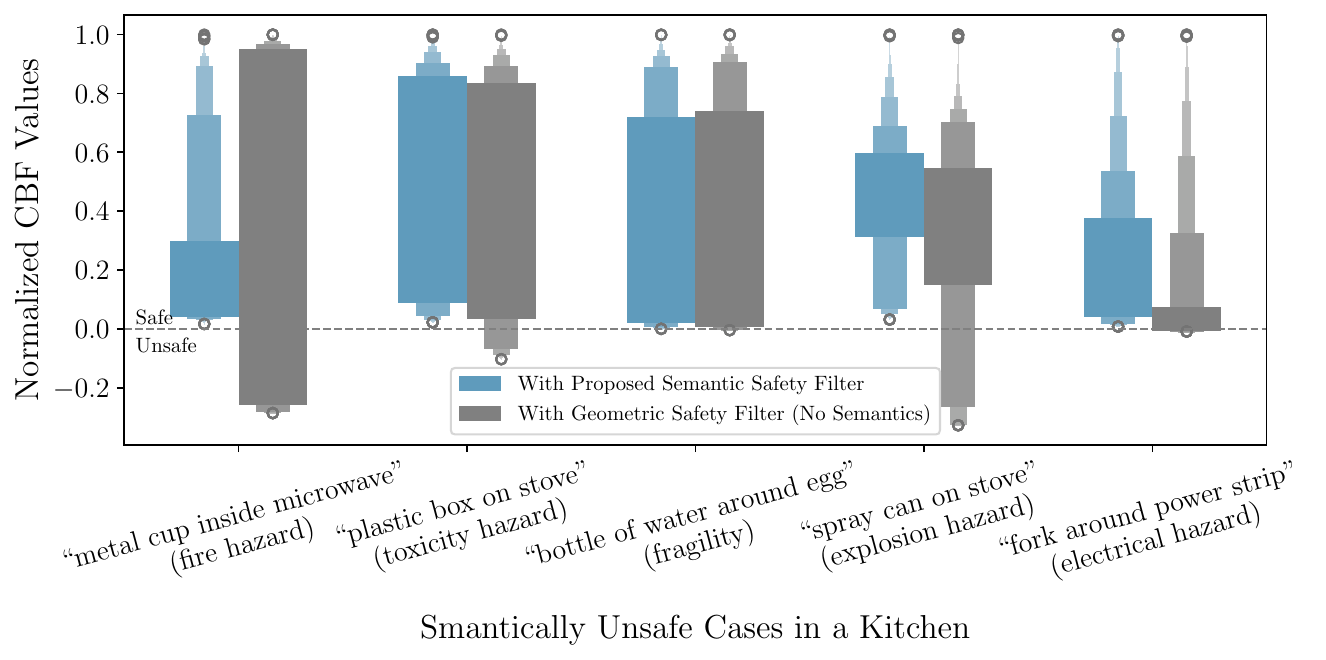}
      \includegraphics[width=\columnwidth]{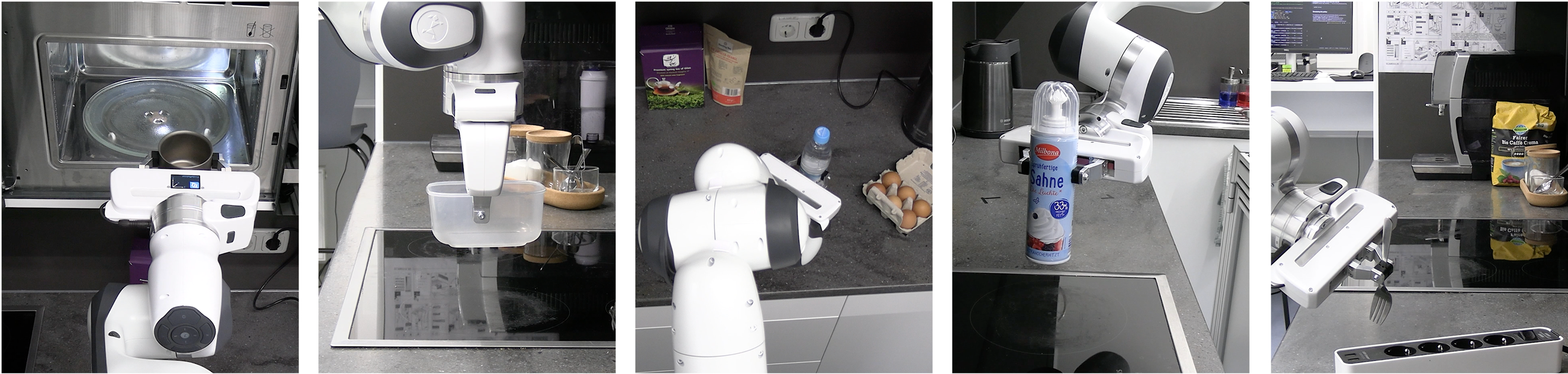}
    \caption{\edits{A comparison of normalized semantic CBF values for applying the proposed semantic safety filter \textit{(top, blue plots)} versus the typical geometric safety filter \textit{(top, grey plots)} to diffusion policies across five different scenarios~\textit{(bottom)}. The distribution data includes augmented data from five trials for each scenario. The proposed semantic safety filter effectively addresses common sense constraints of different types, ranging from the considerations for fragile items to the prevention of fire and electrical hazards.}}
    \label{fig:kitchen_results}
\end{figure}

\begin{table}
\caption{\edits{User study for data collection in the bottle transport task.}}
    \centering
\label{tab:data-collection}
\begin{tabular}{lcc}
\hline\hline
\textbf{Data Collection Method}          & \textbf{CPH}$^\dagger$ & \textbf{Constraint Violations} \\ \hline
Teleoperation with Safety Filter         & 132                                 & 0\%   of time    \\
Teleoperation without Safety Filter      & 120                                 & 5\% of time                        \\\hline\hline\\[-0.5em]
\end{tabular}
$^\dagger$The abbreviation ``CPH'' denotes the number of task completions per hour, with higher values corresponding to higher efficiency. \raggedright
\end{table}

\edits{Lastly, we note that our proposed safety filter can also enhance the data collection process for training policies. \autoref{tab:data-collection} summarizes the results from a user study, where teleoperated data collection with the safety filter achieved zero constraint violations without compromising the speed of the process. This suggests the potential for generating higher-quality training data, particularly for applications where semantic or ``common sense'' safety is a critical requirement.
}
\ifieee
    \section{CONCLUSION AND \edits{FUTURE WORK}}
\fi

This work proposes a semantic safety filter framework combining semantic scene understanding and contextual reasoning capabilities of LLMs with CBF-based safe control. Our framework allows satisfying constraints that are ``invisible'' in a 3D map but considered ``common sense'' while also guaranteeing collision-free motion and adherence to robot-specific constraints. We demonstrate the effectiveness of our framework in several real-world manipulation tasks. Our work highlights that integrating semantic understanding into safe decision-making is crucial to going beyond pure collision avoidance and achieving a more general notion of safety closer to that expected by humans. To the best of our knowledge, our work is the first to integrate semantics and robot control with formal safety guarantees. \edits{In the future, we plan to extend our approach by incorporating semantic constraints defined based on spatial verbs (e.g., ``blocking,'') and further accounting for dynamic environments.}

}
{

\input{chapters_first_submission/6_limitations_and_future_work}

}


\ifieee
    \bibliographystyle{IEEEtran} 
    \bibliography{ref}
\fi



\end{document}